\documentclass[review, 11pt]{elsarticle}

\journal{Pervasive and Mobile Computing}

\usepackage{algorithm}
\usepackage{algorithmic}

\begin{document}

\begin{frontmatter}

\title{Real-time Context-aware Learning System for IoT Applications}

\author{Bhaskar~Das\corref{mycorrespondingauthor}, Jalal~Almhana}
\address{Department of Computer Science, Université de Moncton, Moncton, NB, Canada}
\cortext[mycorrespondingauthor]{Corresponding author}
\ead{das.bhaskar.1981@gmail.com, jalal.almhana@umoncton.ca}

\begin{abstract}
We propose a real-time context-aware learning system along with the architecture that runs on the mobile devices, provide services to the user and manage the IoT devices. In this system, an application running on mobile devices collected data from the sensors, learned about the user-defined context, made predictions in real-time and manage IoT devices accordingly. However, the computational power of the mobile devices makes it challenging to run machine learning algorithms with acceptable accuracy. To solve this issue, some authors have run machine learning algorithms on the server and transmitted the results to the mobile devices. Although the context-aware predictions made by the server are more accurate than their mobile counterpart, it heavily depends on the network connection for the delivery of the results to the devices, which negatively affects real-time context-learning. Therefore, in this work, we describe a context-learning algorithm for mobile devices which is less demanding on the computational resources and maintains the accuracy of the prediction by updating itself from the learning parameters obtained from the server periodically. Experimental results show that the proposed light-weight context-learning algorithm can achieve mean accuracy up to 97.51\% while mean execution time requires only 11ms. 
\end{abstract}


\end{frontmatter}


\section{Introduction}
\label{sintro}

The Internet of Things (IoT) is a vision where everyday devices embedded with computing technology can communicate with one another via the Internet. The IoT devices will play an important role in improving quality of life in many domains such as smart living, transportation, education, agriculture, industry, and the like. By 2020, approximately 212 billion devices will be deployed globally \cite{c2} and will consume 45\% of the internet traffic by 2022 \cite{c1}. IoT-related healthcare industry is expected to grow to \$1.1 - \$2.5 trillion by 2025 globally \cite{c4}, and according to ``Navigant Research", the building automation systems market is expected to reach \$102.0 billion in 2025. Statistics show substantial progress in the field of IoT \cite{c5}. 

Due to the technological advancement of the mobile devices, a new application domain has emerged, called context-aware computing, in which the system can make use of environmental information from collected sensor data and respond accordingly without requiring any user intervention \cite{c6}. Many applications of IoT devices incorporate humans in the model \cite{c7}, which opens new opportunities in many fields including power management, smart homes, healthcare, and the automotive industry. In this paper, we will define the context to detect changes in the environment which can be identified by sensors corresponding to the users. However, detection by processing the sensors' data in the cloud is not suitable for real-time applications that require immediate actions. Various researchers have employed context-aware learning in their works, where they have learned the context of the user and acted automatically based on it. Still, none of the previous research works have offered a solution for real-time context-learning to managee the IoT devices \cite{c8}. In this research, we propose a real-time system that will detect context in real-time with sufficient accuracy. The system is suitable for IoT applications such as health monitoring, home automation, intelligent traffic management systems, environmental monitoring, and smart city management applications, where IoT devices can be managed and connected with the mobile devices. 


Therefore, in this paper, we will extract knowledge using mobile devices in order to make decisions in real time and to store the data on the server for long-term analysis. We will develop algorithms that will learn the context of the user from the data collected through the sensors in real time. Lastly, the algorithm will keep improving the model with the number of usages. 

The main contributions of this paper can be summarized as follows:

\begin{itemize}
\item We propose a real-time context-learning system running on the mobile devices which acquires the data related to the context from sensors and control the IoT devices. The sensors can be internal or external based on the type of device.
\item We have presented two sets of algorithms, of which one utilizes machine learning techniques and the other uses deep learning techniques. 
\item Each set of the context-learning algorithms consists of a light-weight client algorithm and a server algorithm. 
\end{itemize}
 
The rest of the paper is organized in the following manner. Section II presents the literature review while the real-time context-aware learning system architecture is presented in Section III. Section IV describes the learning model followed by a description of our algorithms. The evaluation of the proposed algorithms are presented in Section V, and conclusion is drawn in Section VI. 
 
\section{Literature Review}
\label{slr}

Context-aware computing is a framework through which a system can comprehend the environment automatically and use the information to provide a better service \cite{c9,c11} to the user. 

Nishihara et al. \cite{c12} developed a system consisting of three components: Context Recognition Module, QoS Manager, and Resource Control Module. Although it did not work in real-time, the main point was to maintain a resource table which records the authority of every service along with the essential sensors for the considered contexts.

Donohoo et al. \cite{c13} implemented a framework called AURA which optimizes CPU and screen backlight to conserve power while maintaining a minimum acceptable level of performance. However, this work did not consider multiple context-learning and only focused on user interaction.

Authors in \cite{c20} described that context-aware mobile applications are more profound than distributed and traditionally fixed applications. This work defined a suite of metrics for context-aware mobile applications and demonstrated their impact on the performance and resource utilization in terms of memory, network, and CPU usage.

In \cite{c22}, the authors formulated a learning method based on the context of personal preferences from wearable sensors. With the help of the data acquired from the wearable sensors, the authors designed a user context model which employed machine learning and statistical analysis to learn data online without any external supervision.

Authors in \cite{c23} described a method to reduce the bandwidth usage at the sender by reducing frame rate in mobile video chatting. User experience was unhindered by utilizing internal sensors to drop frames and users adapted to the frame rate. In paper \cite{c24}, authors proposed a context-aware frame rate adaption framework to provide harmony between video quality and bandwidth usage in a mobile video chat application. However, none of these works employed a long-term data analysis technique.

A cloud-assisted lightweight learning framework is proposed by Li et al. \cite{c32}. The solution reduced the data transmission between mobile device and the cloud. The authors utilized the cloud framework to retrain the proposed Convolutional Neural Networks (CNN) in mobile devices from the incremental new training data to increase the accuracy of the model. However, this work did not evaluate the execution time or the power consumption of the model in the mobile devices, which have limited battery life, nor did it examine the response of the model in the event of network failure. 

In \cite{c31}, authors proposed a Viterbi-based context-aware mobile sensing mechanism algorithm. The suggested algorithm employed a trade-off between energy consumption and delay to optimize the sensing schedule of the sensors. Although the solution works in real-time, the authors have not utilized any long-term learning technique to improve the accuracy of the proposed algorithm.

The learning can be performed in the cloud, which offers more computational power and storage space, but users must be connected to the internet to access the service and may have to pay for the cloud service. Learning can also be performed on the mobile device, producing quick results without requiring an internet connection, but is limited by computational power, storage capacity and battery life constraints \cite{c33}. Therefore, to circumvent these shortcomings, in this paper, we propose a real-time context-learning system that can run on mobile devices and is able to upgrade itself from the learning parameters obtained from the cloud. In this paper we will use the term context-aware learning and context-learning interchangeably.

\section{Real-time Context-aware learning system Architecture}
\label{srtcsa}

\begin{figure*}[!ht]
\centering
\includegraphics[width=1\textwidth]{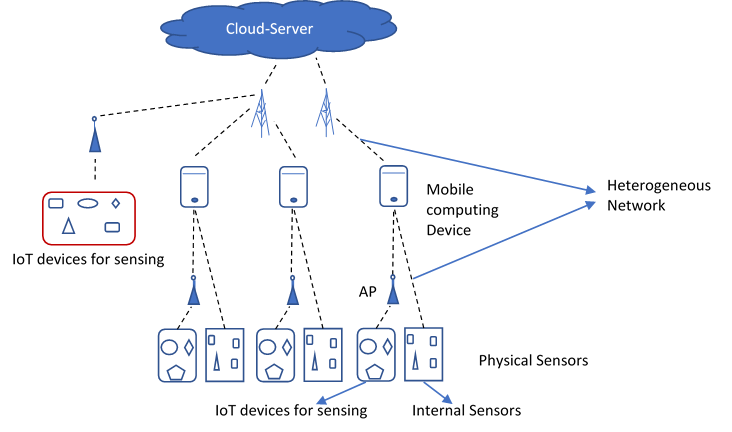}
\caption{Real-time context-aware learning architecture.}
\label{fig_archi}
\end{figure*}

In this paper, we define the context as any information used to understand the environmental changes from the user's perspective, based on sensor data such as location, identity, activity, etc., as proposed in the work of \cite{c34}. Our definition of the context-aware learning system states that the system offers services based on the context, as defined by the user. For example, if a user defined the activity as context and the service required by the user is to play a certain type of music when the user is jogging while another type of music is played when the user is driving, then the context-aware learning system should be able to identify the user-defined context based on the sensor data and provide the required service accordingly by playing the appropriate type of music. Therefore, the context-aware learning system architecture proposed in this section is a general one where users can define the context and services according to their needs and the system will provide appropriate services based on the context. The context-aware learning system requires cloud computing infrastructure which bridges numerous IoT devices to the core computing structure of the cloud. This architecture also requires a light-weight decision-making systems that runs on the mobile devices such as smart-phones, because of the computing-power which currently is not available in most of the IoT devices. Also, mobile devices are already easily available. In this section, we will use the term sensor and sensor node to refer to a physical sensor.

\subsection{Assumptions}
\label{sbasum}

Real-time context-awareness in IoT devices has yet to materialize in a large-scale commercial environment. For this reason, the following assumptions are made in relation to the proposed architecture.

\begin{itemize}
\item Mobile devices, such as smart-phones, are used for the real-time learning by utilizing the data captured from sensors to control and monitor the IoT devices.

\item Mobile devices may have internal sensors.

\item IoT devices can act as an external sensor devices. Though they may have limited computational power, we believe that with advances in technology, the computational power of these devices will increase signicantly and that they will be capable of processing and learning from the captured data \cite{c36, c37}.

\item Internal sensors present in IoT devices can be viewed as autonomous sensors, which can execute applications that affect the sensing and data processing tasks. On the other hand, non-autonomous sensors are external sensors, which cannot process data or make predictions because of lacking computational power. Some sensors may have limited or little computational capabilities, which is not sufficient to execute user defined applications that can influence sensing and data processing tasks, may be called semi autonomous sensors, but in this work they will be referred as non-autonomous sensors.

\item The sensing unit contains control units, sensor delay, sleep interval and power module. 

\item Light-weight learning and decision-making processes are performed in the mobile devices and the computationally heavy learning tasks are performed in the cloud. 

\item Mobile devices store sensor data temporarily and transmit it to the cloud infrastructure for permanent storage.

\item IoT devices can be mobile or stationary. 

            
\end{itemize}

\subsection{System outline}
\label{sbso}

A generic real-time context-learning system architecture can be thought of as a system consisting of three tiers, as shown in Fig. \ref{fig_archi}.

\begin{itemize}
\item Tier 1: The sensor nodes reside at the bottom layer. They are responsible for sensing the environment, which helps to determine the context, and they transmit the raw data to the mobile devices and the cloud server.

\item Tier 2: It consists of the mobile devices which are responsible for receiving data from the sensor nodes, receiving learning parameters from the cloud server, transmitting control information to the sensor nodes, transmitting data to the cloud server and control the functioning of the IoT devices to provide better services to the users.

\item Tier 3:  The cloud server layer is the top layer of the proposed architecture. It consists of multiple servers which run the learning algorithm and provide the learning parameters to the mobile devices. The data-centers, connected with the servers, are responsible for storing the sensor data generated from the numerous sensors. 
\end{itemize}

\subsection{Architecture details}
\label{sbad}

\begin{figure*}[!ht]
\centering
\includegraphics[width=1\textwidth]{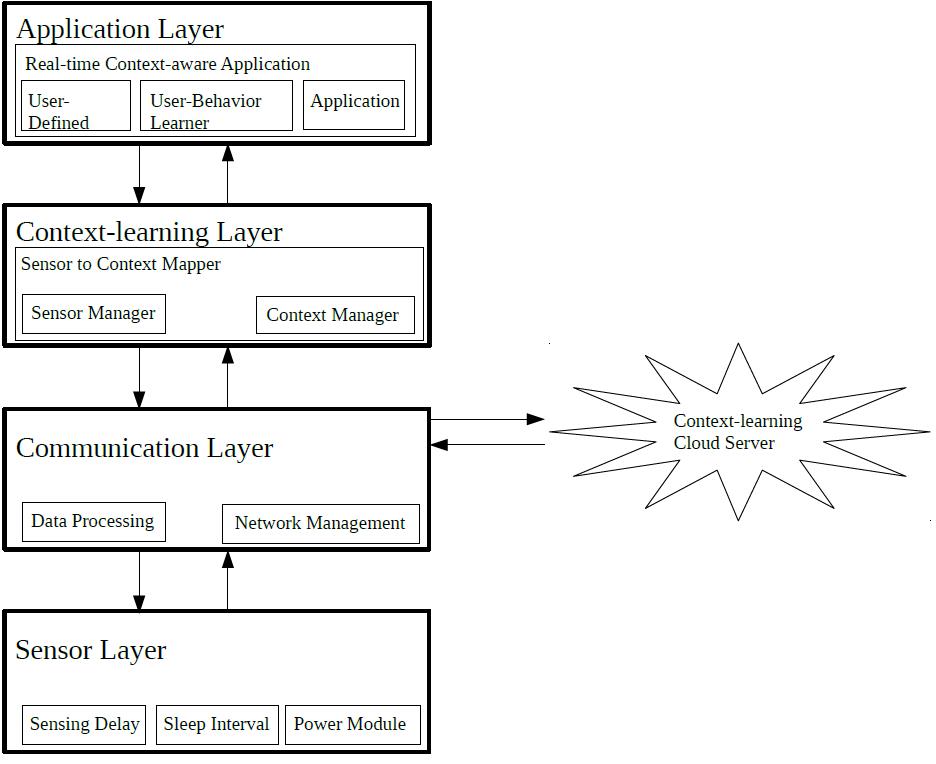}
\caption{Detailed architecture of the real-time context-aware learning system.}
\label{fig_darchi}
\end{figure*}

Fig. \ref{fig_darchi} presents the detailed architecture of the proposed real-time context-aware learning system. First layer is the Sensor Layer, consists of the sensor nodes which can be independent sensors connected to the mobile devices or cloud server through the wireless or wired communication links. Sensors can also be internal and situated within the mobile devices. Sensor nodes transmit data to the mobile devices at a specified interval. The sensing interval is instructed by the sensor nodes on the cloud server depending upon who is receiving the data. The Communication layer connects the sensors to the sensor manager and facilitates data transmission from the sensor and controls information from the sensor manager. The Sensor manager instructs the sensors on how to gather data by setting the sensor delay and the sleep interval. The Sensor delay determines the rate at which a sensor node collects sensed data. The Sleep interval is the time interval for a sensor node to refrain from its sensing tasks. The Power module manages the power consumption of sensor nodes. The Sensor delay, the sleep interval and the power module can take instructions from mobile devices and the cloud server for optimum performances.

Second layer is the Communication Layer which is responsible for transmitting data from the sensors to the mobile devices, and from mobile devices to the cloud server. This layer has two functions: data processing and network management. The data processing unit processes the data before transmitting it to the mobile devices or the cloud server. Meanwhile, the Network management unit manages the communication processes among various heterogeneous networks. It also implements priority-wise data transmission when two or more sensor nodes want to transfer data at the same time.

The Context-Learning Layer is in the third layer of our architecture, which consists of the sensor to context mapper sub-layer that further contains sensor manager and the context manager. The sensor to context mapper sub-layer access sensors through sensor manager and map a virtual sensor to a context. It is also responsible for creating a virtual sensor from the physical sensors. A virtual sensor may contain more than one physical sensor.  The sensor manager unit collects data from the sensors and controls the properties of sensor nodes by setting sensing delay, sensor priority, and sensor sleep interval. The context manager maintains the context priority along with the corresponding sensors required by each context.

The fourth layer is the Application Layer, which contains the context-aware user behavior learner, real-time context-aware application, and other third-party applications. A user can select a context which is previously defined from the available set. The real-time context-aware application unit can be used globally on all applications or work with a particular application. Our model generates contexts based on the available sensors. The Context-Aware user BehAvior Learner (CABAL) monitors the interaction pattern of the user with the device and how it is affected by environmental factors which can be represented by the data or obtained from the sensors. Therefore, CABAL evaluates user interaction pattern changes according to the sensor value change and sends it to the Real-Time Context-aware Learning Application (RTCLA). The RTCLA executes the contexts according to their overall priority on all applications. However, the user can omit some applications, which will not be affected by the RTCLA.

The cloud server can receive sensor data from mobile devices or directly from the sensor nodes and transmit the control instruction to the sensor nodes with the help of the communication layer. The learning parameters are sent to the mobile devices periodically to update the real-time learning system. The cloud server may contain data-centers and cloud storage to store the sensor data and the learning parameters. The learning unit of the cloud server updates the learning parameters from the sensor data collected over time. The data processing unit updates the corresponding data processing unit present in the communication layer while the sensor manager unit manages the sensors.

\section{Learning Model}
\label{ssm}

\begin{figure}[ht]
\centering
\includegraphics[width=0.6\textwidth]{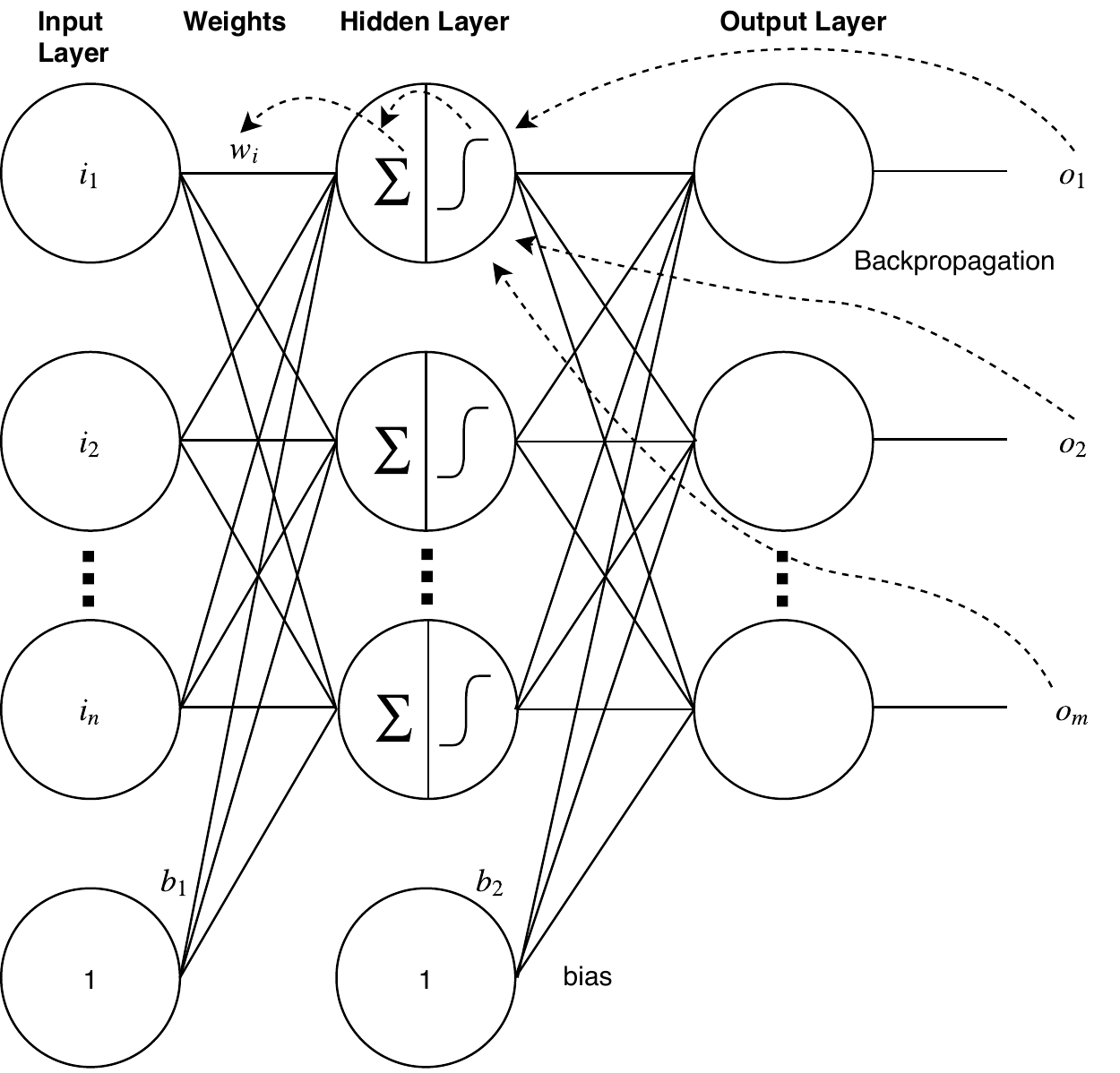}
\caption{Neural Network.}
\label{fig_nn}
\end{figure}

An interconnected group of nodes, which are modeled after neurons in human brains, are used to build an artificial neural network \cite{c25}. The artificial neural network can be used to model non-linear and complex relationships among the inputs and outputs, can infer unseen relationships within the data which is helpful to create a generalized model that is able to predict unseen data, can work with a wide range of inputs, and generally performs better when compared to other types of algorithms. The deep learning systems performed better with complex input and output mapping. A large number of deep learning systems are based on a neural network because of its feature learning ability when compared to other approaches such as decision trees. Therefore, considering the advantages of neural network based systems, it is considered it in this paper. Generally, neural networks consist of layers which are made up of nodes. Our neural network, illustrated in Fig. \ref{fig_nn}, has three layers; input, hidden and output layers, with each consisting of nodes. The nodes of different layers are connected by links, which are associated with weights. In the forward propagation process, the network takes input data from the sensors, processes them through the hidden layer and sends the result to the output layer. A context may be identified using several sensors. In such a case, each sensor data should be used as an input value in the network. The total input of a hidden layer is determined using the following equation as presented in \cite{c25}.

\begin{equation}
\label{eq1}
\eta_{h_\lambda}=b_l+\sum_{j=1}^{\omega}w_j\ast\sum_{k=1}^{\iota}{i_k\ }
\end{equation}

Where $\eta_{h_\lambda}$ is the input of the hidden node $h_\lambda$, $w_j$ is the weight for the link between the input node  $i_k$ and hidden node $h_\lambda$. The bias is denoted as $b_l$. There is a total of $\iota$ input nodes and $\Lambda$ hidden nodes in the network. Therefore, each hidden node will use equation \ref{eq1} to determine the total input. The output of a hidden node is obtained by the activation function as demonstrated in \cite{c25}, which is provided below.

\begin{equation}
\label{eq2}
\sigma_{h_\lambda}=\ \frac{1}{1+\ e^{-\eta_{h_\lambda}}}
\end{equation}

Where $\sigma_{h_\lambda}$ denotes the output of the hidden layer $h_\lambda$. We have used the sigmoid function in this paper for activation. After obtaining the output from the hidden nodes, we measure the error for each output using the following function, which is also called squared error function as mentioned in \cite{c26}. 

\begin{equation}
\label{eq3}
E= \sum_{x=1}^{m}\frac{1}{2}\left(a_x-o_x\right)^2 
\end{equation}

Where $a_x$ is the actual output for output node $x$ and $o_x$ is the predicted output. In the backward propagation process \cite{c26}, we update the weights and the biases of the network by comparing them with the actual output values. For the output layer, we will use the following equation, which is explained in \cite{c26}.

\begin{equation}
\label{eq4}
\frac{\delta E}{\delta w_{h_\lambda o_x}}=\ \frac{\delta E}{\delta\sigma_{o_x}}\ast\ \frac{\delta\sigma_{o_x}}{\delta\eta_{o_x}}\ast\ \frac{\delta\eta_{o_x}}{\delta w_{h_\lambda o_x}}
\end{equation}

Where $\frac{\delta E}{\delta w_{h_\lambda o_x}}$ is the gradient with respect to the link between output node $o_x$ and the hidden node $h_\lambda$ and is obtained using the chain rule \cite{c27}. For the hidden layer, the equation from \cite{c26} is used.

\begin{equation}
\label{eq5}
\frac{\delta E}{\delta w_{i_kh_\lambda}}= \frac{\delta E}{\delta\sigma_{h_j}}\ast\ \frac{\delta\sigma_{h_\lambda}}{\delta\eta_{h_\lambda}}\ast\ \frac{\delta\eta_{h_j}}{\delta w_{i_kh_\lambda}}
\end{equation}

Where $\frac{\delta E}{\delta w_{i_kh_\lambda}}$ is the gradient with respect to the link between input node $i_k$ and the hidden node $h_\lambda$. The weights are updated using the following equation derived from \cite{c26}.

\begin{equation}
\label{eq6}
\varphi_{w_y}= w_y - \Gamma * \frac{\delta E}{{\delta w}_y}
\end{equation}

Where $\varphi_{w_y}$ is the value of the updated weight, $w_y$ is the previous weight, $\Gamma$  is learning parameter, and $\frac{\delta E}{{\delta w}_y}$ is the effect of the weight on the total error. The network will update all the weights using equation \ref{eq6}.

\subsection{Algorithms}
\label{sbalgo}

The real-time system proposed in this paper consists of two types of context-learning algorithms. One algorithm runs on the server, called the server-side algorithm. The server-side context-learning algorithm is an independent algorithm which receives the sensor data from the mobile devices and learns the context. The other algorithm is the client-side algorithm which runs on the mobile devices. These algorithms do not require high computational power and take less time to learn the context. They depend on the server, which provides the learning parameters, to maintain accuracy in real-time. The client-side algorithm is responsible for transmitting the sensor data to the server and offload the computationally heavy learning processes to the server. In the server, we have implemented two types of context-learning algorithms; one with machine learning, denoted as Context-Learning (CL), and another with deep learning, denoted as Deep Context-Learning (DCL).

\begin{algorithm}[H]
\label{algo1}
 \caption{Client-side light-weight context-learning Algorithm using Deep Context-learning (ADCL)}
 \begin{algorithmic}[1]
 \renewcommand{\algorithmicrequire}{\textbf{Input:}}
 \renewcommand{\algorithmicensure}{\textbf{Output:}}
 \REQUIRE Get the sensor data $S = \left\lbrace  s1, s2, \cdots, s_\iota \right\rbrace $
 \ENSURE  Context $C = \left\lbrace c1, c2, \cdots, c_x \right\rbrace $
 \STATE Network $\gets$ Get network from the server
 \STATE $C \gets$ predict($Network, S$)
 \\ \
 \STATE function predict($Network, S$)
 \STATE $ \Gamma \gets S$
 \FORALL {node in each layer in the Network}
 \STATE {Activation-value: calculated from the equation \ref{eq1}}
 \STATE {Output: calculated from the equation \ref{eq2}}
 \STATE {$\Gamma$: output of hidden layers become input for output layer}
 \ENDFOR
 \STATE Return $C$ from index value of the highest values of the $\Gamma$
 \STATE end function
 \end{algorithmic}
 \end{algorithm}

Algorithm \ref{algo1} describes the working of the client-side light-weight context-learning algorithm using deep learning, denoted as Algorithm for Deep Context-Learning (ADCL). The client-side ADCL works with the server-side DCL algorithm. DCL receives the context along with the corresponding set of sensors from the client mobile devices. Service providers can facilitate the context-learning through cloud services. In this paper, we present a generic system framework for context-learning that can be implemented by the users according to their requirements. In any case, the server-side algorithm collects data from the sensor, learns the context by employing a deep learning technique using a backpropagation algorithm of the artificial neural network, and transmits the updated weights to the ADCL, which is on the client device. The ADCL predicts the context from the learning parameters for the current set of sensor data. The predict function described in the Algorithm \ref{algo1} is the same as the forward pass function used in the backpropagation algorithm in DCL.

\begin{algorithm}[H]
\label{algo2}
 \caption{Client-side Light-weight Context-learning (LCL) algorithm using machine learning}
 \begin{algorithmic}[1]
 \renewcommand{\algorithmicrequire}{\textbf{Input:}}
 \renewcommand{\algorithmicensure}{\textbf{Output:}}
 \REQUIRE Get the sensor data $S = \left\lbrace s1, s2, \cdots, s_\iota \right\rbrace $ 
 \ENSURE  Context $C =  \left\lbrace c1, c2, \cdots, c_x \right\rbrace $
 \STATE $W \gets$ Get weights from the server, where $W = \left\lbrace w1, w2, \cdots, w_\omega \right\rbrace $
 \STATE $\tau$ $\gets$ Get threshold value from the server, where $\tau = \left\lbrace t1, t2, cdots, t_x \right\rbrace $
 \STATE $C \gets$ predict ($W, S, \tau$)
 \\ \ 
 \STATE function predict($W, S, \tau$)
 \STATE $\Gamma$ $\gets S$
 \STATE $Z \gets$ Get $Z$ from equation \ref{eq1}
 \STATE $P \gets$ Calculate the $P$ using equation \ref{eq2}
 \STATE $C \gets$ Predict the context by comparing output value with the $\tau$
 \STATE Return $C$
 \STATE end function
 \end{algorithmic}
 \end{algorithm}

Algorithm \ref{algo2} describes the working of the client-side, denoted as Light-weight Context-Learning (LCL) algorithm using machine learning. LCL works with its server-side counterpart algorithm CL, which employed machine learning techniques. CL receives data from the sensors to identify a predetermined context. CL then sends the learning parameters, which are weights of the network along with the threshold values, to the LCL. LCL then uses these learning parameters to predict the context for the current set of sensor data.

Both client-side algorithms mentioned in Algorithm \ref{algo1} and \ref{algo2} use the learning parameters provided by their server-side counterparts to predict the context in real-time. However, the connection between the mobile devices and the server may not be consistent due to several factors such as mobility, disasters, hardware failure and the like. In those circumstances, it is not possible for the server to transmit the learning parameters. The client uses previously transmitted learning parameters to predict the context, which may reduce the accuracy, but provide the result in real-time. We illustrate in the Evaluation section that the average precision of the proposed algorithm is quite high as the learning is performed in real-time. According to the research work presented in \cite{c30}, humans react within ~150-200ms. Therefore, we demonstrate in the Evaluation section that our algorithms predict within 150ms.

\section{Evaluation}
\label{seval}

In this section, we analyze the performance of the proposed real-time system. We have evaluated the system’s accuracy and execution time using an Android application. We implemented two light-weight context-learning algorithms, LCL and ADCL, as described in the previous section. These two algorithms are specially tailored to run on mobile devices with a limited computational capability and learn the context in real-time. Subsequently, it is easier to make a decision autonomously in real-time. Therefore, these algorithms can be applied by the user in diverse types of applications according to their requirements. 

\subsection{Experimental Setup}
\label{sbes}

We will evaluate the four algorithms: CL, LCL, DCL, and ADCL. Our objective is to learn the context in the least amount of time from the sensor data with sufficient accuracy. We have implemented all four context-learning algorithms on the Nexus 5x smartphone.  The algorithms run on Android Marshmallow version 6.0, API level 23. To evaluate the proposed algorithms, we have used various datasets as well as implemented an example which learns whether the user is still or in motion from the accelerometer data. The smartphone connects to the Internet using Wi-Fi to send sensor data to the server. We created our neural network with $\varphi$ hidden layers, $\iota$ input layers and $\varrho$ output layers, such that $\vartheta = \vert (\iota + \varrho)/2 \vert $. We have presented six datasets, iris \cite{c28}, wheat seeds \cite{c29}, heart disease diagnosis \cite{c40}, spoken Arabic digit \cite{c39}, high-quality recordings of Australian sign language signs \cite{c38}, and road condition data \cite{c35} to evaluate the performance of our algorithms. We have presented accuracy obtained by the proposed ADCL algorithm in the results and discussion sub-section with six different datasets to illustrate that our algorithm can work with multiple contextual data and produce good results in real-time. We know that the activity recognition experiment presented in this paper is a trivial one but we have used for the evaluation purpose of the proposed system which include our overall architecture, which includes learning from the sensor data in real-time, storing the data in the server and updating the real-time learning algorithm by the learning parameters obtained from the cloud server. However, the rest of the experiments with the five datasets along with the vehicular dataset are non-trivial and used to show the efficiency of the real-time learning algorithm. Table \ref{tes} describes the list of parameters used in the present experiment, along with their associated values.

\begin{table}[!t]
\label{tes}
\centering
\caption{Experimental Setup}
\begin{tabular}{l|p{5.5cm}}
\hline
\textbf{Parameter} & \textbf{Value}\\
\hline
Smartphone	& Nexus 5x\\
Android version	& Marshmallow version 6.0\\
Android API level &	23\\
Sensors	& Accelerometer\\
Internet Connection	& Wi-Fi\\
Datasets for validation & iris \cite{c28}, wheat seeds \cite{c29}, heart disease diagnosis \cite{c40}, spoken Arabic digit \cite{c39}, high-quality recordings of Australian sign language signs \cite{c38}, and road condition data \cite{c35}\\
Server-side context-learning & Python based solution\\
Neural Networks (ADCL) &  3 layers (Input, Hidden and Output)\\
Learning Rate (ADCL) &  0.3\\
Each neuron's weight (ADCL) & Initialized with random numbers in the range of 0 to 1\\
bias & Initialized with random numbers in the range of 0 to 1\\
Transfer neuron activation & using sigmoid function\\
\hline
\end{tabular}
\end{table}

\subsection{Results and Discussion}
\label{sbrnd}

\begin{figure}[ht]
\centering
\includegraphics[width=0.75\textwidth]{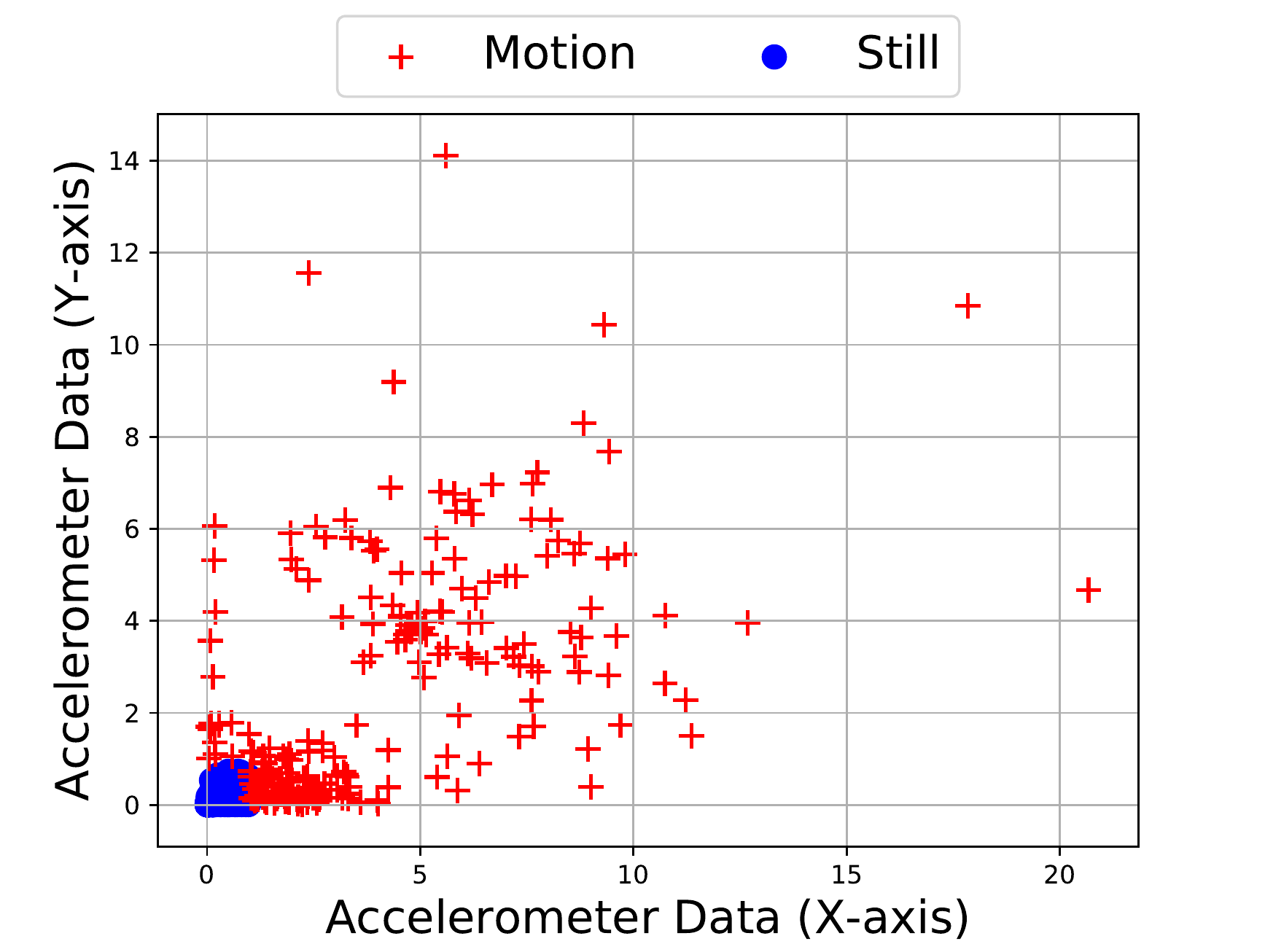}
\caption{Snapshot of Accelerometer Data.}
\label{fig_datasnapshot}
\end{figure}

Fig. \ref{fig_datasnapshot} presents a snapshot of the accelerometer data. In this experiment, we learn the context of a user based on the stillness of the device. Our system collects the accelerometer data automatically and stores it after each usage session as determined by the user. We have considered x-axis and y-axis values of the accelerometer data while leaving the z-axis values. We have observed that the z-axis value of the accelerometer is not useful for our experiment. Therefore, we have only used x and y-axis values of the accelerometer data in this experiment. To train our system, we asked the user to manually mark the cases when they are using the smartphone while moving and when they are in a still position. The blue circle represents the scenarios when the user is still while using the smartphone whereas the red plus denotes the cases when the user is in motion. One can observe from Fig. \ref{fig_datasnapshot} that when the user is still, the accelerometer readings are lower than in the cases where the user is in motion.

\begin{figure}[ht]
\centering
\includegraphics[width=0.75\textwidth]{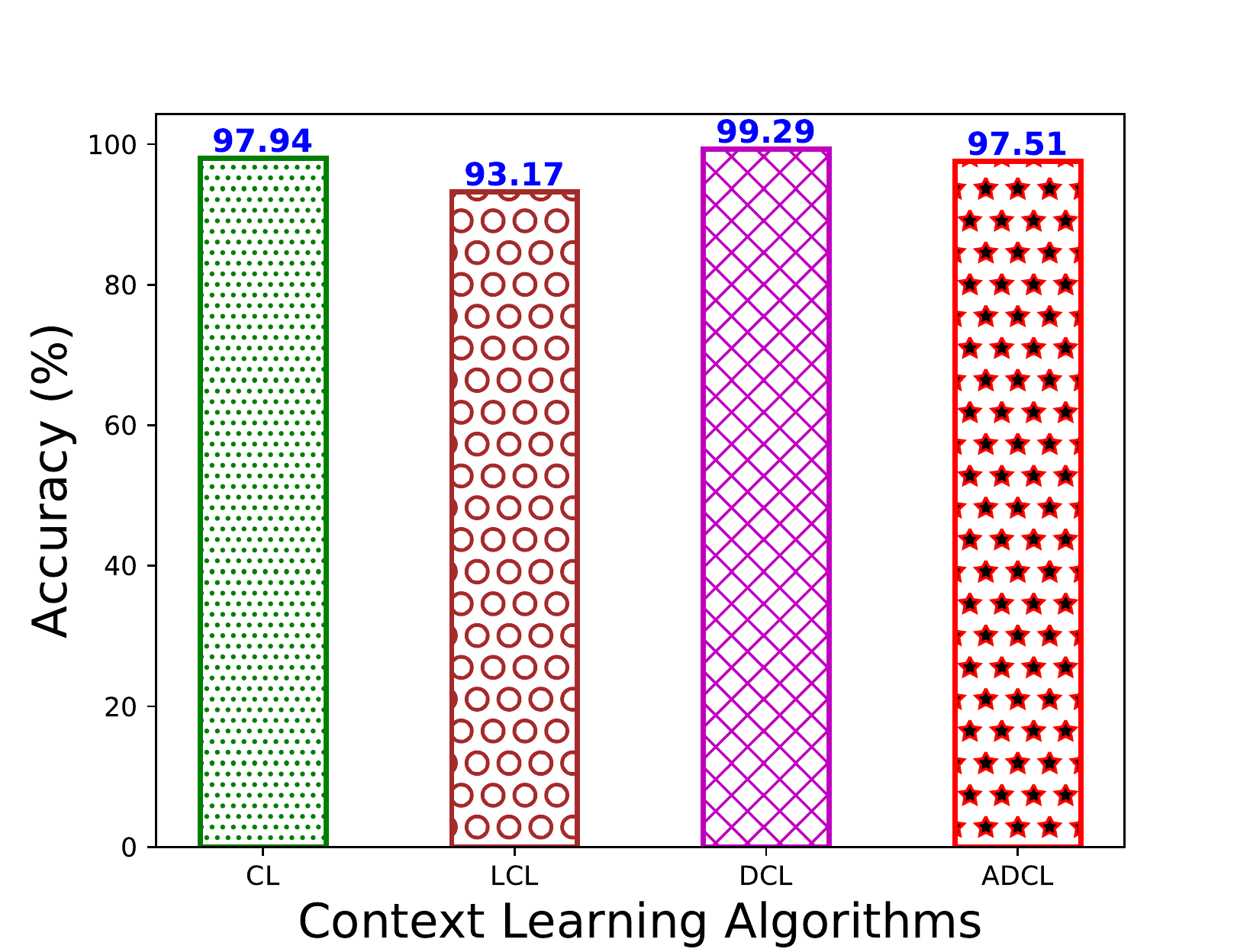}
\caption{Average accuracy comparison of Context-learning Algorithms.}
\label{fig_accuracy}
\end{figure}

Fig. \ref{fig_accuracy} shows the mean accuracy of each context-learning algorithm. The DLC algorithm, which runs at the server, achieves the highest mean accuracy (99.29\%). The CL algorithm, executing at the server, gives the second-best result at 97.94\%. The ADCL algorithm (average accuracy 97.51\%) which runs at the smartphone closely follows the CL server algorithm. The LCL algorithm, which runs on the smartphone, produced the lowest score (93.17\%). Hence, it is a wise choice to run the deep learning algorithm at the server and use its light-weight version to make the prediction in real-time on the mobile devices in order to achieve the most accurate results.

\begin{figure}[ht]
\centering
\includegraphics[width=0.75\textwidth]{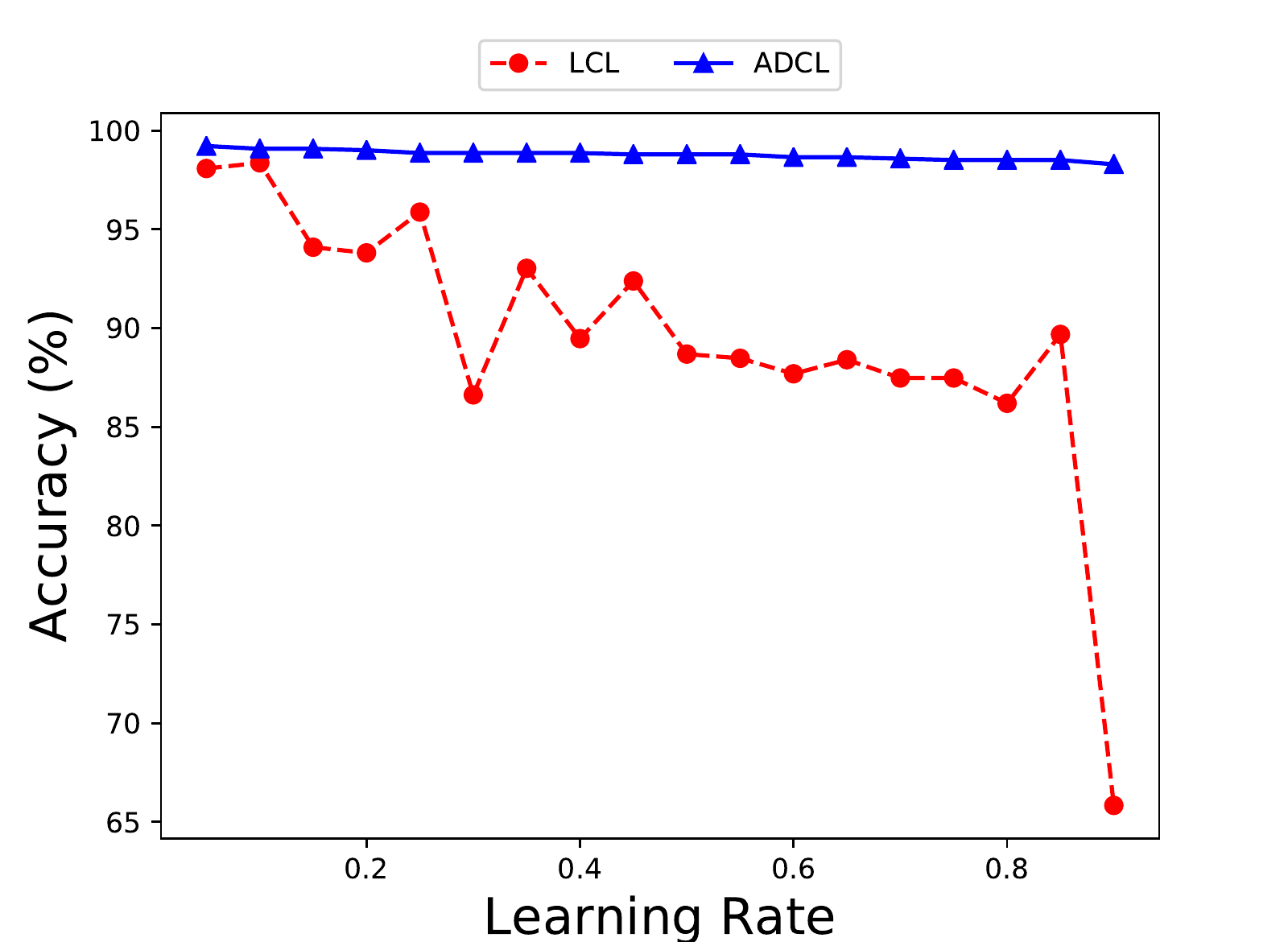}
\caption{Effect of varying learning rate on the Accuracy.}
\label{fig_lrate}
\end{figure}

Fig. \ref{fig_lrate} demonstrates the effect of a varying learning rate for LCL and ADCL. In ADCL, the lower learning rate produces higher accuracy. Although it is also the same for the LCL algorithm, it does not follow a leaner trend. If we use a lower learning rate, then we should train the model more, which is suitable for server-side algorithms. Therefore, we have selected a low learning rate for the server algorithm, which is 0.05, and a relatively higher learning rate of 0.3 for the client-side algorithm.

\begin{figure}[ht]
\centering
\includegraphics[width=0.9\textwidth]{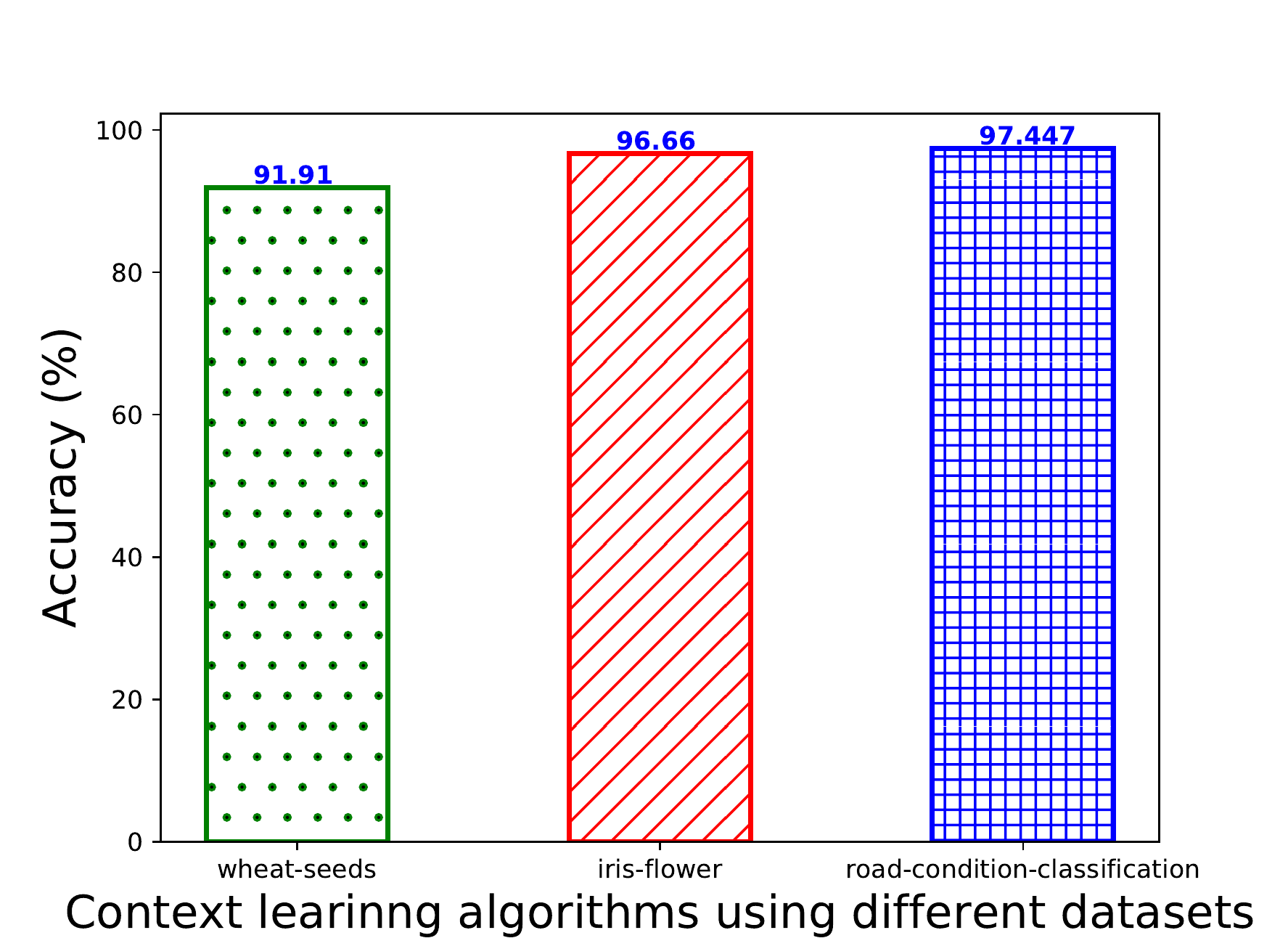}
\caption{Accuracy of ADLC algorithm while using iris datasets, seeds datasets, and road condition data.}
\label{fig_datasets}
\end{figure}

To compare the efficiency of the proposed ADCL algorithm, we have compared it with two standard machine learning datasets in Fig. \ref{fig_datasets}. These are the wheat seeds dataset \cite{c29} denoted as “wheat-seeds”, and the iris flower dataset \cite{c28} denoted as “iris-flower”. The wheat seeds dataset comprises seven input variables and one output variable that decides the type of wheat out of three species. The iris flower dataset involves prediction of the flower from four input variables. The output variable can be divided into three classes. The accuracy of these two datasets is illustrated in Fig. \ref{fig_datasets}. The proposed ADCL algorithm obtained 96.66\% accuracy for the iris flower dataset, followed by 91.91\% for the wheat seeds dataset. We have also compared our ADCL algorithm with another context-learning algorithm presented in \cite{c35}, where the context is the condition of the road which was learned from the accelerometer readings obtained from the vehicle. The authors classified the road conditions, based on the accelerometer readings, into three types, they are Smooth, Average, and Rough. The experimental results showed that our algorithm increases the accuracy by 8.777\%, when compared to the results obtained in \cite{c35}.

\begin{figure}[ht]
\centering
\includegraphics[width=1\textwidth]{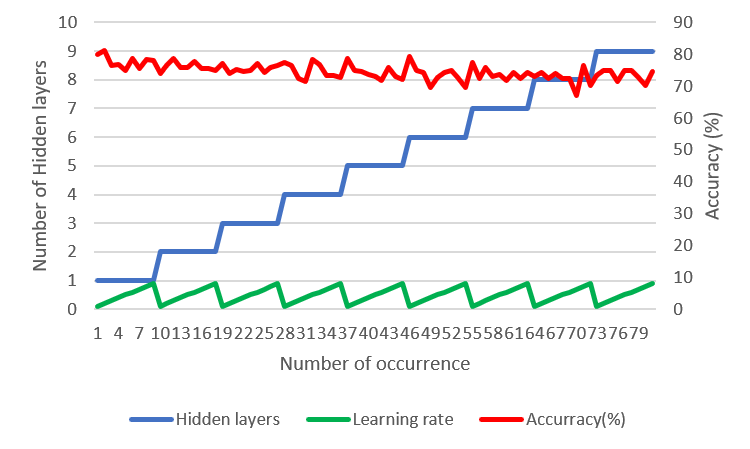}
\caption{Heart Disease dataset.}
\label{fig_heartdisease}
\end{figure}

We have illustrated the efficiency of our proposed ADCL algorithm with the heart disease dataset \cite{c40} in Fig. \ref{fig_heartdisease}. The x-axis of the figure represents the number of occurrence, which is the averaged results of our experiment presented for each learning rate. The y-axis of the figure presents the number of hidden layers. We have considered maximum of nine hidden layers starting from one and for each hidden layer we have presented the learning rate from 0.1 to 0.9. For this heart disease dataset we have acquired the highest accuracy of $81.33\%$ at 0.2 learning rate with only one hidden layer. We have used 14 attributes dataset of the Cleveland database and the output field, which refers to the presence of heart disease in the patient valued from 0 that represents no heart disease to 4.

\begin{figure}[ht]
\centering
\includegraphics[width=1\textwidth]{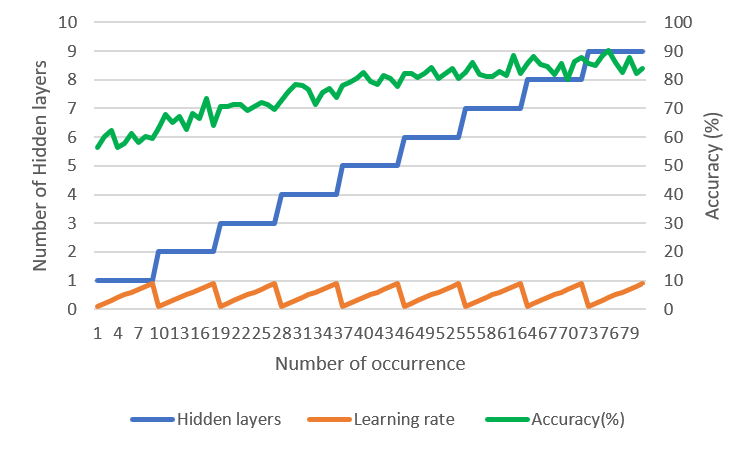}
\caption{Spoken Arabic digit dataset.}
\label{fig_arabsign}
\end{figure}

We have illustrated the efficiency of our proposed ADCL algorithm with the spoken Arabic digit dataset \cite{c39} in Fig. \ref{fig_arabsign}. This dataset consists of time series data of 13 Frequency Cepstral Coefficients (MFCCs), which were recorded from 44 males and 44 females, where native Arabic speakers between the ages 18 and 40 spoken 0 to 9, total ten, Arabic digits. The best result we obtained in terms of accuracy is $90.29\%$ at $0.4$ learning rate with 9 hidden layers.

\begin{figure}[ht]
\centering
\includegraphics[width=1\textwidth]{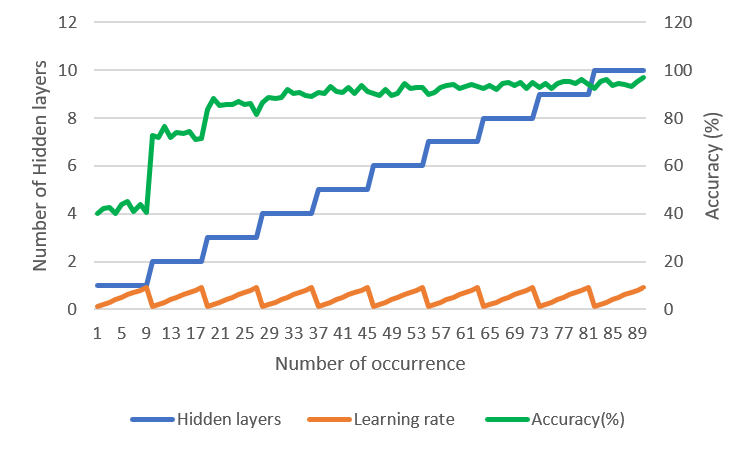}
\caption{Australian sign language signs (High Quality) dataset.}
\label{fig_aus}
\end{figure}

We have illustrated the efficiency of our proposed ADCL algorithm with the Australian sign language signs (High Quality) dataset \cite{c38} in Fig. \ref{fig_aus}. To capture the data two Fifth Dimension Technologies (5DT) gloves, one right and one left, along with two Ascension Flock-of-Birds magnetic position trackers, one attached to each hand are used and stored in a Intel Pentium II 266MHz PC with 128MB RAM. The data ware collected from a native Auslan signer over a period of nine weeks. Total 2565 signs were collected of which each sign contains 27 samples. The best result we obtained in terms of accuracy is $97.11\%$ at $0.9$ learning rate with 10 hidden layers.

\begin{figure}[ht]
\centering
\includegraphics[width=1\textwidth]{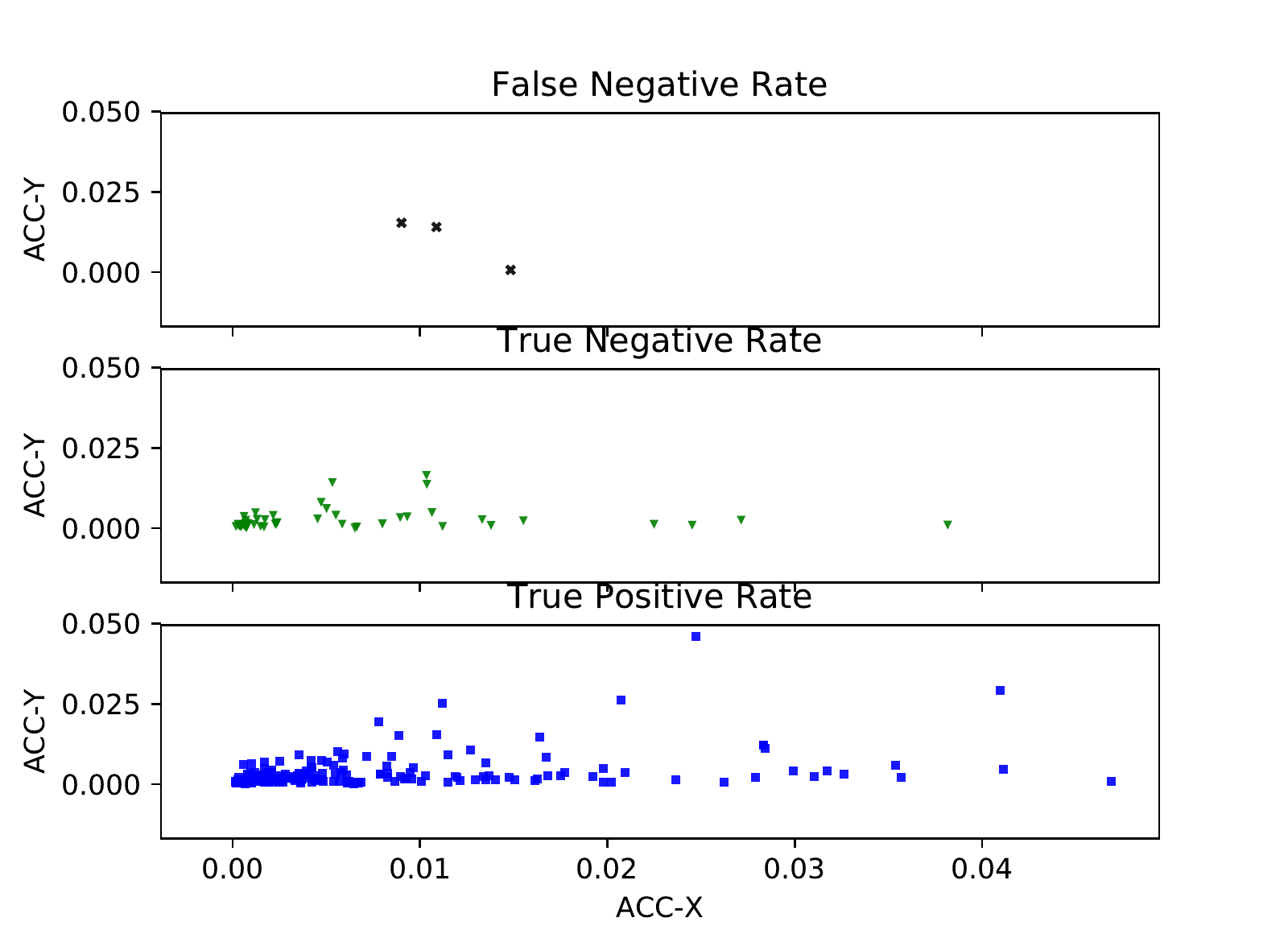}
\caption{True positive, true negative and false negative rate.}
\label{fig_taccuracy}
\end{figure}

In Fig. \ref{fig_taccuracy}, we illustrate the snapshot of true positive, true negative and false negative rates for accelerometer data while using the ADCL algorithm. In the figure, the true positive rate is denoted by blue squares, the true negative rate is denoted by green inverted triangles and the false negative rate is denoted by black crosses. In the figure, the x-axis presents the data reading from the accelerometer (x-axis values) and the y-axis of the plot represents the y-axis reading of the accelerometer data. The occurrence of the false positive rate is very minimal; therefore, it is not presented in the figure. In this experiment, the percentage of true positive, true negative and false negative values are 13.70\%, 84.88\%, and 1.42\% respectively. We have achieved the mean accuracy of 98.505\% for this experiment.

\begin{figure}[ht]
\centering
\includegraphics[width=0.8\textwidth]{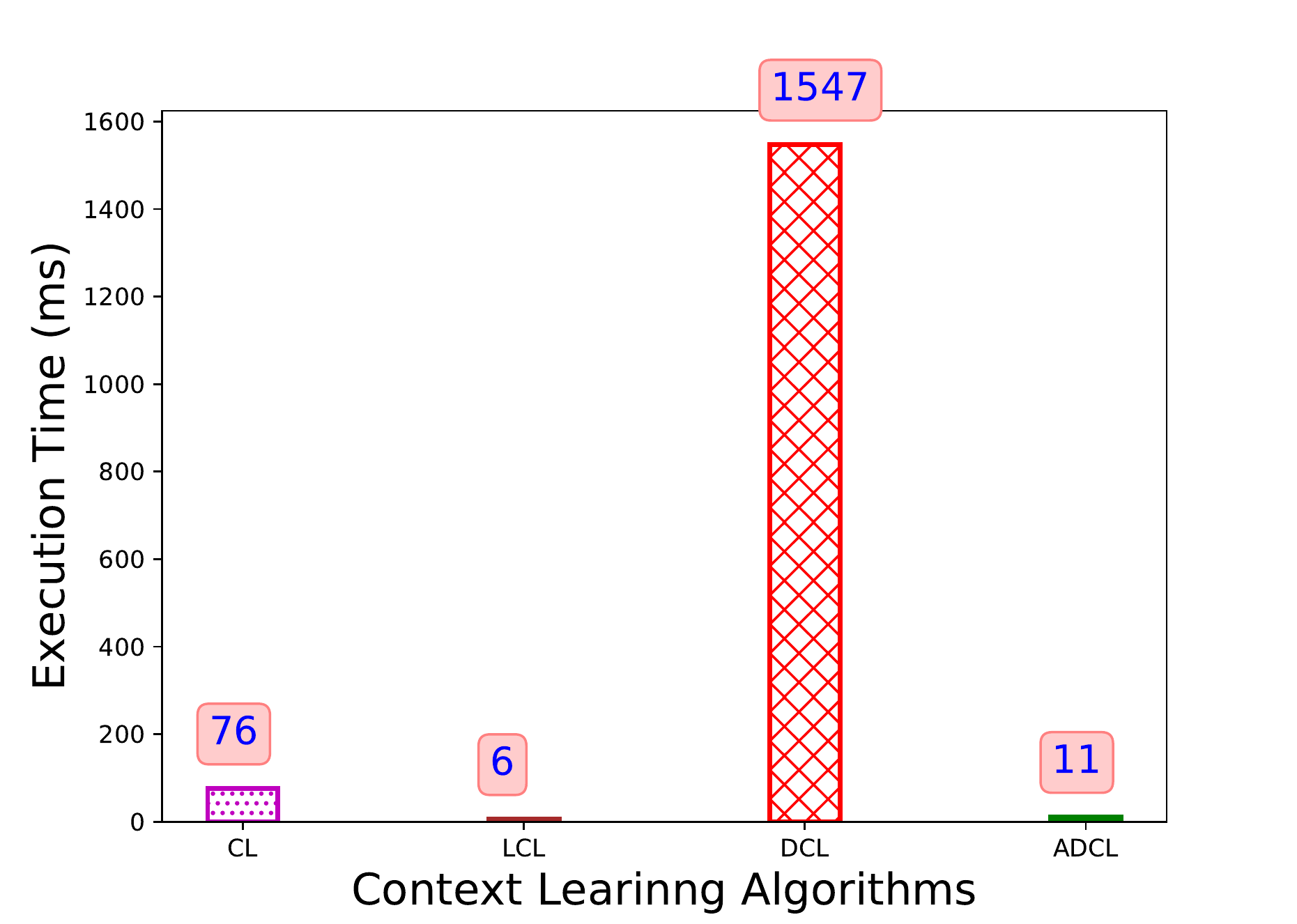}
\caption{Execution Time of context-learning algorithms in milliseconds.}
\label{fig_execution}
\end{figure}

We have compared the execution of the four algorithms for the activity recognition experiment. All algorithms are implemented on the Android for comparison. One can observe from Fig. \ref{fig_execution} that LCL takes the least amount of time  to execute (6ms), followed by ADCL which takes 11ms. CL takes 76ms while DCL needs 1.5s to run on an Android device. By comparing these results to Fig. \ref{fig_darchi}, we can establish that ADCL produces more accurate results in the least amount of time. Hence, ADCL is most suitable for real-time context-learning systems. Therefore, we recommend it be used in mobile devices for context-learning. However, one may use CL on mobile devices with better computational power when the accuracy of the learning is the priority. Hence, it is up to the user to choose which algorithm to use based on the situation.

\section{Conclusion}
\label{scon}

In this paper, we present a context-learning system for the IoT applications where the cloud server assists the learning process in the mobile devices which can manage the IoT devices to provide better service to the users. We have proposed an architecture that is made with four interconnected components which are sensors, IoT devices, mobile devices and cloud server, where mobile devices acquired sensing data from sensors, transmit the data to the cloud server for long-term analysis, receives learning parameters from the cloud server and manage the IoT devices according to the context of the user and provide solution in real-time. We focused on the learning of user-defined context in real-time in mobile devices with limited computational capabilities while providing better accuracy. A machine learning algorithm is not suitable for mobile devices since it adversely affects the accuracy and the time needed to learn the context. Therefore, to solve this issue, this paper puts forward a light-weight algorithm suitable for mobile devices. The proposed system consists of a client algorithm and a server algorithm. The server algorithm receives the sensor data directly from the sensors or other intermediate devices, learns the context and transmits the learning parameters back to the mobile devices. The client algorithm, which runs on the mobile devices, uses these learning parameters to learn the context in real-time. Experiments are conducted by implementing the proposed algorithms on an Android application in order to demonstrate how the execution time affects the accuracy. In addition, we have used different standard datasets to evaluate the efficiency of the two proposed light-weight learning algorithms in terms of accuracy, one based on machine learning techniques and the other on deep learning techniques. In summary, the current study focuses on the real-time learning of the user-defined context from the sensor data using light-weight learning algorithms that run on mobile devices and provide accurate results by receiving learning parameters from the server.


\bibliography{ref}



\end{document}